\documentclass{ieeeaccess}
\usepackage{cite}
\usepackage{amsmath,amssymb,amsfonts}
\usepackage{algorithmic}
\usepackage{graphicx}
\usepackage{textcomp}

\usepackage{multicol}
\usepackage{multirow}
\def\BibTeX{{\rm B\kern-.05em{\sc i\kern-.025em b}\kern-.08em
    T\kern-.1667em\lower.7ex\hbox{E}\kern-.125emX}}
\begin{document}
\history{Date of publication xxxx 00, 0000, date of current version xxxx 00, 0000.}
\doi{*}

\title{Centerline Depth World for Reinforcement Learning-based Left Atrial Appendage Orifice Localization}
\author{\uppercase{Walid Abdullah Al}\authorrefmark{1},
\uppercase{Il Dong Yun\authorrefmark{2} \IEEEmembership{Member, IEEE}, and Eun Ju Chun}.\authorrefmark{3}}
\address[1]{Department of Computer and Electronic Systems Engineering, Hankuk University of Foreign Studies, Yongin, 17035 South Korea (e-mail:walidabdullah@hufs.ac.kr)}
\address[2]{Department of Computer and Electronic Systems Engineering, Hankuk University of Foreign Studies, Yongin, 17035 South Korea (e-mail:yun@hufs.ac.kr)}
\address[3]{Department of Radiology, Seoul National University Bundang Hospital, Seongnam, 13620 South Korea (e-mail:humandr@snubh.org)}
\tfootnote{This research was supported by Basic Science Research Program through the National Research Foundation of Korea (NRF), funded by the Ministry of Education, Science, Technology (No. 2019R1A2C1085113).
This work was also supported by grant No. 02-2018-043 from the Seoul National University Bundang Hospital (SNUBH) Research Fund.}

\markboth
{Abdullah Al \headeretal: Centerline Depth World for Reinforcement Learning-based Left Atrial Appendage Orifice Localization}
{Al \headeretal: Centerline Depth World for Reinforcement Learning-based Left Atrial Appendage Orifice Localization}

\corresp{Corresponding author: Il Dong Yun (e-mail: yun@ hufs.ac.kr) and Eun Ju Chun (humandr@snubh.org).}

\begin{abstract}
Left atrial appendage (LAA) closure (LAAC) is a minimally invasive implant-based method to prevent cardiovascular stroke in patients with non-valvular atrial fibrillation. Assessing the LAA orifice in preoperative CT angiography plays a crucial role in choosing an appropriate LAAC implant size and a proper C-arm angulation. However, accurate orifice localization is hard because of the high anatomic variation of LAA, and unclear position and orientation of the orifice in available CT views.
Deep localization models also yield high error in localizing the orifice in CT image because of the tiny structure of orifice compared to the vastness of CT image.
In this paper, we propose a centerline depth-based reinforcement learning (RL) world for effective orifice localization in a small search space. In our scheme, an RL agent observes the centerline-to-surface distance and navigates through the LAA centerline to localize the orifice. Thus, the search space is significantly reduced facilitating improved localization. The proposed formulation could result in high localization accuracy comparing to the expert-annotations in $98$ CT images. Moreover, the localization process takes about $8$ seconds which is $18$ times more efficient than the existing method. Therefore, this can be a useful aid to physicians during the preprocedural planning of LAAC.
\end{abstract}


\titlepgskip=-15pt

\maketitle

\section{Introduction}
\label{sec:introduction}

\begin{figure*}[!ht]
\centering
\includegraphics[scale=1.0]{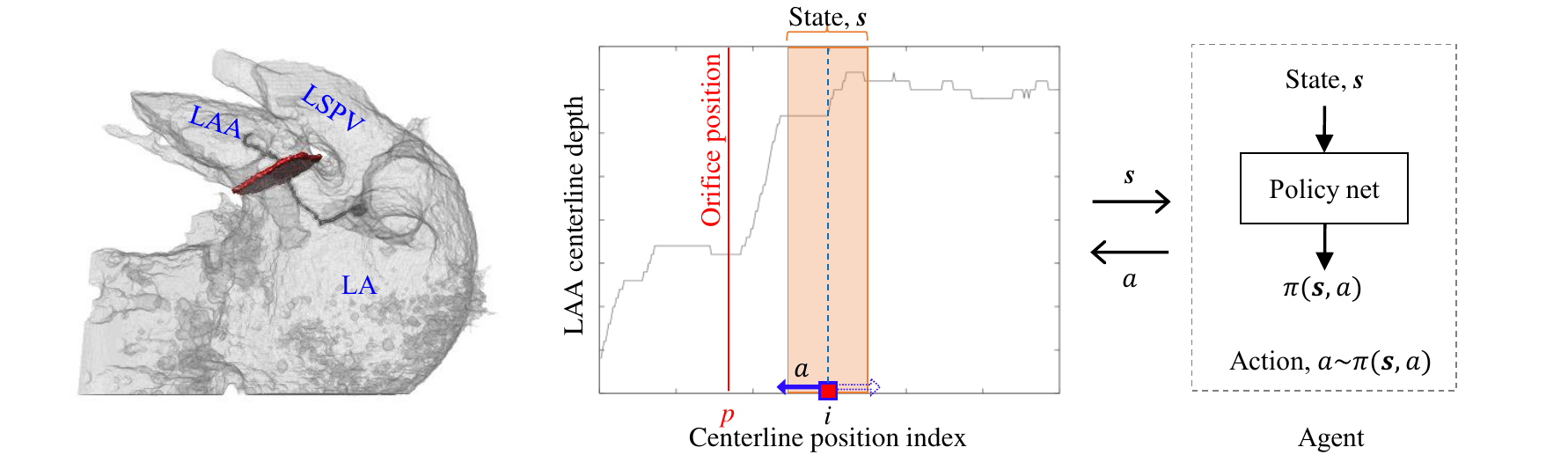}

\caption{{\bf The proposed centerline depth world for LAA orifice localization.} (Left) centerline from LAA to LA, and the orifice (red region) are shown in the cropped input volume. (Right) RL world for the corresponding centerline. Red square mark indicates the localization agent position $i$. Observing the local centerline depth state $s$, the agent moves forward or backward along the centerline to converge at the orifice position $p$ indicated by the red vertical line. Agent decides the optimal moving action $a$ based on its policy $\pi(\boldsymbol{s},a)$.}
\label{fig:cdworld}
\end{figure*}

\PARstart{A}{trial} fibrillation (AF) characterized by rapid and irregular atrial beating is one of the most common heart abnormalities \cite{afmunger2014}. According to the report by Zoni-Berisso et al. \cite{zoni2014epidemiology}, AF affects
$2$--$3$\% of the population, with a high increase comparing to $0.3$--$1$\% in $2005$ \cite{fuster2015american}. One of the most prevalent
sites of thrombus formation leading to AF-associating cardiovascular stroke is the left atrial appendage (LAA) \cite{burrell2013usefulness}. To treat high-risk AF-patients, a minimally invasive implant-based strategy called LAA closure (LAAC) is widely used \cite{reddy2013left}. In this strategy, a closure device is implanted at the appendage orifice  to prevent the thrombus from entering into the bloodstream. For preoperative interventional planning, LAA assessment deserves duteous attention where the major concern to the physicians during the assessment is the detection of the orifice. Accurate identification of the orifice plays a significant role because it contributes to choosing the occlusion device of an appropriate size and obtaining a proper C-arm angulation for intervention.
\par
Left atrial appendage can be viewed as an anatomical projection from the main left atrial (LA) chamber, and the LAA orifice (or, ostium) is the narrow opening of the appendage to the atrium (Fig.~\ref{fig:cdworld}). Preoperative planning of LAAC utilizes CT angiography to acquire 3D measurements during the orifice assessment. However, localizing the appendage orifice is a challenging task due to significant variation in shape, size, and orientation of the appendage. Moreover, conventional CT views do not allow a clear understanding of the appendage anatomy. In addition, the appendage and the left atrium share a common intensity with no separation wall against each other. Consequently, accurate orifice localization becomes laborious and time-consuming for the physicians. Therefore, a computational method for appendage orifice localization can be useful to the physicians by reducing their effort and speeding up the preoperative planning.

With the available LAA orifice assessment works mainly limited to manual approaches \cite{budge2008analysis,wang2010left,  walker2012anatomical}, the major research-focus is on the LAA segmentation where no indication is provided for the orifice. Prior appendage segmentation works are generally semi-automatic requiring a manual volume-of-interest (VOI) annotation. In most of these works, the primary goal is to segment the left atrium, where appendage is included in the segmentation result without any distinction \cite{tobon2013left, liu2017left}. Zheng et al. \cite{zheng2012precise, zheng2014multi} suggests segmenting the left atrium by separate shape-constrained segmentation of different parts of the LA including the appendage. Later, the gap between the main LA chamber and the appendage is filled by projecting the proximal ring of appendage towards the main chamber. Because the orifice usually resides in this gap, such smooth projection may risk oversimplifying the irregular orifice shape.

There are a number of works treating LAA segmentation as the major objective. Grasland-Mongarin et al. \cite{grasland2010combination} also presented shape-constrained deformable models to segment the cardiac chambers, where LAA was extracted using mesh inflation. However, model-based approaches face problem in coping with the varying LAA structure. Model-free approaches usually have two steps: (i) axial slice-wise 2D segmentation proposals (using parametric max-flow \cite{wang2016left} or fully convolutional networks \cite{jin2018left}), and (ii) 3D refinement (using 3D conditional random field \cite{jin2018left}). Such model-free approaches could cope better with the varied anatomy of LAA. Nonetheless, all these segmentation approaches include a part of the left atrium in their segmented results without presenting a clear separation for the orifice.

Leventi{\'c} et al. \cite{leventic2017semi} separated the appendage from the left atrium in their region growing-based segmentation. The separation plane is obtained from three additional points manually marked in the segmented volume, supposedly representing the LAA orifice. Recently, they extended their work to compute the orifice location following a semi-automatic approach requiring a threshold to obtain initial mask and a seed-point in LAA \cite{leventic2019left}. To the best of our knowledge, this is the only existing computational method dealing with orifice localization. 

The orifice detection approach of Leventi{\'c} et al. \cite{leventic2019left} is based on scanning the cross-sectional area along the centerline. Because cross-sections inside the LA continues to be larger after the orifice, this method identifies the largest continuous rise in the cross-sectional areas. The orifice location is detected at the closest local minima before such rise. The optimal cross-section for each point in the centerline is exhaustively determined by taking the minimum area from multiple cross-sectional planes tilting the plane by up to 40 degrees. Overall, the whole process become computationally expensive taking about 2.5 minutes \cite{leventic2019left}. 

Most importantly, using such a fixed rule for localizing the orifice may not be a robust solution considering the highly varied anatomy of the appendage. The unavoidable noise and artifacts in CT can also become potential threats against such localization rule. False local minima may arise posing threat to identifying both the largest uninterrupted rise inside LA and the closest local minima before.

Machine learning approaches showed improvements in several medical image analysis tasks under such high variation \cite{zheng2012automatic, al2018automatic}. With the advent of deep learning, several end-to-end localization models have been proposed. Such deep models can predict location of keypoints or landmarks of interest directly from image. Convolutional neural network (CNN)-based cascaded regression network \cite{lv2017deep, kwak2020facial} is a widely used architecture to regress the direct location. For anatomical landmark localization in medical images, spatially regressing Gaussian heatmaps \cite{payer2016regressing, payer2019integrating} around the target location showed fruitful results. However, training such end-to-end models to localize the orifice from raw image is difficult because of the tiny orifice structure compared to the vast search space of 3D CT image.

\par
In this work, we aim at ensuring robust LAA orifice localization under the high anatomical variation. Instead of setting a fixed rule \cite{leventic2019left} for detecting the orifice, we suggest a learning-based approach to utilize the rich context in expert-annotated data. In our approach, a reinforcement learning (RL) agent is trained to navigate through LAA centerline to localize the orifice. For effective learning, we propose a centerline depth-based RL environment, which has a small search space to enable easier learning, yet showing strong potential for discriminating the orifice location. Due to the training difficulty using raw image, we equip the agent with the centerline-to-surface distance information for simpler but better representation about the anatomical progression. Fig.~\ref{fig:cdworld} illustrates the proposed RL world for orifice localization. For centerline seed initialization, the proposed semi-automatic method requires a simple click roughly at the LAA tip, which is clearly visible in the orthogonal CT views.

For improved evaluation, we use a dataset significantly larger than the dataset used in the previous orifice localization work \cite{leventic2019left}. Our dataset also includes real patients who have undergone the LAAC procedure. Besides comparing the orifice center location similar to the previous study \cite{leventic2019left}, we also compare the orifice plane and area against the expert orifice annotations for a fruitful evaluation of the detected orifice. We show that the proposed localization method yields significantly lower detection error compared with the rule-based \cite{leventic2019left} and the other learning-based solutions. 
Taking only $8$ seconds, the proposed orifice localization method can be a useful guide to the physicians during preoperative planning of LAAC. 

\section{Methodology}
\label{sec:method}
To accelerate the proprocedural assessment of LAA orifice in CT angiography, we suggest a robust orifice localization approach formulating the localization process as a reward-maximization problem of a RL agent.  Using the LAA centerline as the search space for localization, we propose a simple RL world called the centerline depth world. The localization agent in such a world interacts by moving one step at a time along the centerline, to eventually converge near the orifice position after a number of steps. 
\par
Centerline depth refers to the distance of centerline from the nearest surface-point. Usually, anatomy of the left atrial appendage suggests a gradually increased cross-sectional area from the LAA orifice to the LA unlike the area from LAA tip to the orifice \cite{leventic2019left}. Therefore, the alteration of centerline depth from the LAA to LA provides powerful information for the agent to localize the orifice. Localizing the tiny orifice in the vast CT image using an end-to-end deep localization model \cite{lv2017deep, payer2016regressing} is difficult. Instead, the centerline can serve as a small but useful search space to track the anatomy. Such centerline-based search strategy is also used in the previous orifice localization study \cite{leventic2019left} utilizing the above anatomical progression towards the atrium. The previous study computed the minimal cross-sectional area at each centerline point to search for the closest local minima before the largest uninterrupted increase. However, the desired cross-sectional areas along the centerline is highly noisy and computationally expensive. The depth computation, on the other hand, is highly efficient and has relatively low noise, while providing an equivalent representation (shown later in Fig.~\ref{fig:cases}). 

In our RL world, the agent observes the local depth-map and navigates through the centerline to finally reach the orifice. We first describe the procedure of computing the centerline from a given CT image, then move to presenting the proposed centerline depth world for orifice detection.

\subsection{LAA Centerline Extraction}  
This process requires a manually annotated initial seed roughly at the LAA tip. Marking this initial seed is convenient and fast because the process is not sensitive to the seed position as long as it resides inside the LAA tip before the LAA neck. From the initial seed, we obtain the LAA centerline in three steps. Firstly, we binarize the CT image extracting the organs-of-interest (OOI) from the background. Secondly, we perform the Euclidean distance transform from the binarized image. Finally, we grow the centerline from LAA to LA using the resultant distance map. 

Let us denote the initial LAA seed by $\boldsymbol{x}_s$. The OOI extraction should be extensive in nature so that the resultant region includes a significant part of the left atrium beyond the appendage orifice. Otherwise, the following detection using the axis-to-surface distance can be problematic due to insufficient data related to the progression after the orifice. Existing LAA segmentation techniques could be used for appendage extraction. However, as discussed in the previous section, the model-based approaches among those has the problem of coping with structural variation \cite{jin2018left}. The supervised slice-wise mask prediction with 3D CRF-based refinement comes with a large computational cost \cite{jin2018left, wang2016left}. Supervised approaches also requires laboriously marking the ground truth for 3D segmentation mask for all the training volumes. 

\begin{figure}
\includegraphics[scale=1.0]{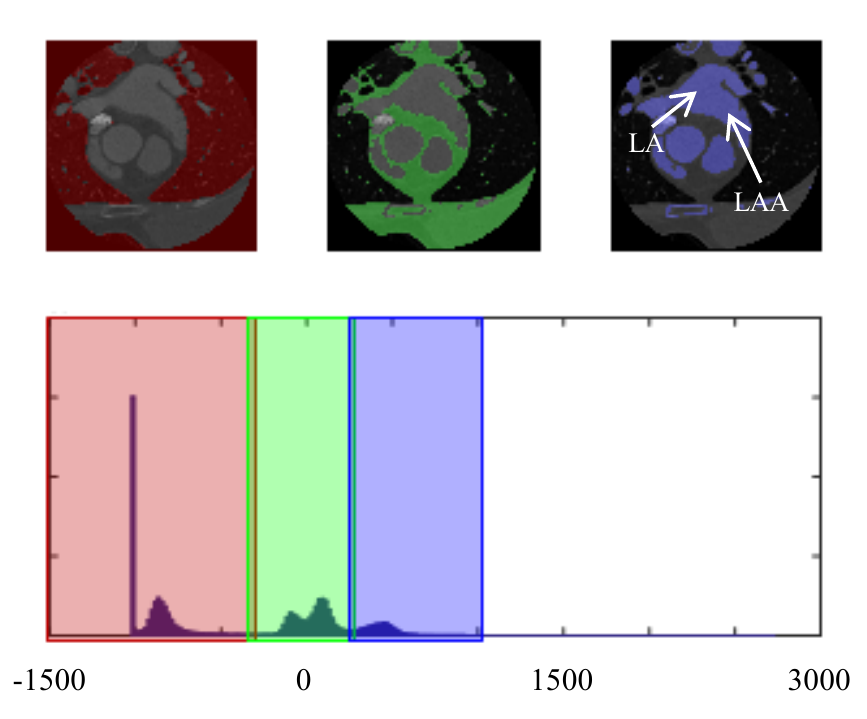}
\caption{{\bf Intensity distribution of a CT image (in Hounsfield unit)}. Intensity-based geodesic distance thresholding can extract the LA and LAA regions. }
\label{fig:intensity_distribution}
\end{figure}

In contrast, we adopt a model-free 3D geodesic distance based unsupervised approach. The LA and LAA has an even and easily distinguishable intensity distribution compared with the background (Fig.~\ref{fig:intensity_distribution}). Therefore, we can simply obtain the desired LAA and LA regions based on the geodesic distance from the LAA seed.
Obtaining the desired region directly from 3D geodesic distance instead of 2D slice-wise prediction enables faster computation. Geodesic distance-based extraction also allows the seed to grow where it finds similar intensity without any constraint. Therefore, the resultant region can cover the entire appendage following the intensity including the anatomical irregularities. This advantage is also discussed in \cite{jin2018left}. The major problem of the this approach is reported to be the inclusion of the adjacent left superior pulmonary vein (LSPV) \cite{jin2018left} because small leakage may occasionally exist from LAA to LSPV. However, this does not affect our centerline depth-based localization. This is because depth is calculated as the centerline distance from the nearest surface (or background) and LAA surface is closer than the LSPV surface. 

\par
First, we obtain a fixed-size VOI relative to the seed location, so that it sufficiently encloses the appendage and a part of the LA. This may also include the left superior pulmonary vein (LSPV). Inside the VOI, we compute the geodesic distance for all the voxels with respect to the localized seed $\boldsymbol{x}_s$. Finally, we can extract the extended LAA by thresholding the distance.
\par  
Initially, we grow the seed by a radius of $5$ voxels to obtain a small seed region $\Omega$. 
For the reference voxel-set in $\Omega$, the geodesic distance at any position $\boldsymbol{x}$ indicates the distance of the shortest geodesic path, which is defined as follows:
\begin{equation}
\begin{split}
\mathcal{D}(\boldsymbol{x}; \Omega) &= \min_{\boldsymbol{y}\in\Omega}{\min_{\zeta \in \mathcal{P}_{\boldsymbol{x},\boldsymbol{y}}} \sum_{\boldsymbol{p}\in \zeta} \sqrt{\big(  |\nabla_{\zeta} \boldsymbol{p}|^2+\alpha|\nabla_{\zeta}I(\boldsymbol{p})|^2\big)}}
\end{split}
\label{eq:transform}
\end{equation}
where $\zeta$ iterates over all paths from $\boldsymbol{x}$ to $\boldsymbol{y}$, $\mathcal{P}_{\boldsymbol{x},\boldsymbol{y}}$, and $\nabla_{\zeta}$ indicates the gradient following a path $\zeta$. $I$ denotes the intensity. $\alpha$ sets the contribution of image gradient over spatial distance. In our experiment, we simply use $\alpha=1$. Such geodesic distance is also used by Criminisi et al. \cite{criminisi2008geos} for segmentation. As opposed to the usual segmentation labeling decided by the comparative closeness to a foreground or background seed, a distance threshold parameter $\lambda$ decides the segmentation in our case of a single foreground seed. In our experiment, a fixed threshold value $\lambda=0.3$ is used for thresholding the normalized geodesic distance. As we showed in Fig.~\ref{fig:intensity_distribution}, the intensity distribution of our desired region is apparently different from the background. Therefore, the threshold value does not have to be precise. We obtained the above threshold value by visual assessment through a trial-and-error process.   
\par
We compute the geodesic distance following the raster scan update scheme used by \cite{criminisi2008geos}. Two sets of forward and backward updates are performed. In the forward update, we visit and update each voxel using \textit{upper-lower slice, top-bottom, left-right} scanning. Backward scan follows the opposite direction. The current voxel is updated using the geodesic distance information of its neighbours that have been visited in the current scan.
\par

After obtaining the binarized CT image with foreground including the LA, LAA, etc., we compute the Euclidean distance transform \cite{maurer2003linear} to have the distance of each voxel from its nearest background. Note that, this is equivalent to the voxel-wise distance to the nearest surface of the extracted region. Let us denote this distance map by $\hat{D}$ . We call this the \textit{depth} of a voxel with respect to the surface.

We grow the centerline by tracking the maximum depth voxels from the initial LAA seed $\boldsymbol{x}_s$ to the LA. At each step $i$, we decide the next centerline voxel $\boldsymbol{x}_{i+1}$ by finding the local maxima among the unvisited neighboring voxels.Thus, we represent the centerline by all of the sequentially tracked voxels as: $\mathcal{M} =(\boldsymbol{x}_{i})_{i=1}^T$. Here, $T$ is the fixed length of the centerline. In our experiment, $T=300$ was sufficient to grow the desired centerline.

The resultant centerline $\mathcal{M}$ serves as the search space for the orifice detection. Anatomy of the left atrial appendage suggests a gradually increased cross-sectional area from the LAA orifice to the LA unlike the area from LAA tip to the orifice. Instead of computing the cross-sectional area about the computed axial points in $\mathcal{M}$, we use the axis-to-surface distance (i.e., the distance of the axial points to the nearest surface) because it renders equivalent information. Moreover, the axis-to-surface distance about an axial point $\boldsymbol{x}_i \in \mathcal{M}$ can be obtained directly from the earlier Euclidean distance transform map $\hat{\mathcal{D}}(\boldsymbol{x}_i)$ without any further computation.

\subsection{Centerline Depth World}
The proposed RL world is defined based on the computed centerline $\mathcal{M}$ and the depth along the centerline voxels. We denote the centerline depth by a vector $\boldsymbol{\hat{d}} = (\hat{d}_i)_{i=1}^{T}$ where $\hat{d}_i = \hat{\mathcal{D}}(\boldsymbol{x}_i)$ is the depth of the $i$-th voxel in the centerline. Fig.~\ref{fig:cdworld} illustrates the centerline depth world. Note that, the agent position in this world suggests the index to a centerline voxel in the actual CT image. We define the target orifice position $p$ in this world to be the index to the closest centerline voxel comparing with the expert annotated orifice center. Initiating from any position index $i$, the goal of the RL agent is to localize this orifice voxel index $p$ in the centerline.

The previous study applied a fixed rule to track the closest local minima before the largest uninterrupted cross-sectional increase (supposedly occurring in the LA after the orifice). This can be sensitive to the highly varying LAA anatomy. For example, there are some cases where there are no local minima at the LAA neck because LAA neck is larger than the tip. Additionally, false local minima may arise due to the high noise in CT.

To handle such issues, we train an agent to optimize its policy for the expert-annotated data containing varied LAA images. The proposed world is a simple 1D world, yet capable of providing discriminative knowledge for the orifice localization agent. The agent initialized at any position index $i$ in the centerline, observes the local centerline depth, and decides an action to move forward or backward along the centerline. Thus taking an episode of steps along the centerline, it eventually converges to the target $p$-th position in the centerline. 

Relating the key elements of RL formulation, state $\boldsymbol{s}$ can be referred to as the local centerline depth map centered at the agent position. Therefore, for any position $i$, the state $\boldsymbol{s}_i = (d_j)_{j=i-k}^{i+k}$  for a state-length of $2k+1$. Observing this state, the agent decides to move forward or backward along the centerline to reach the orifice in the future. Therefore, the action-space can be denoted as: $A=\{forward, backward\}$. The policy $\pi(\boldsymbol{s},a)$ represents the agent-behavior model that outputs the optimal action probabilities $P(a); a \in A$ for a given state $\boldsymbol{s}$. We denote agent-transition by $(i,a,i')$ indicating a transition from position $i$ to $i'$ on the centerline by action $a$.

For training the policy, a discrete reward signal $R$ is fed for each transition, to encourage moves that draws the agent closer to the target or discourage otherwise. We also encourage a reward of convergence when the agent is at a small distance away from the target. Similar rewarding scheme is also showed to be effective in previous RL-based anatomical landmark localization using full image space \cite{al2018partial}. For a transition $(i,a,i')$,  we present our reward function as follows:
\begin{equation}
R(i, a, i') = 
\begin{cases}
+2, & \text{ if } |p-i| \le \tau\\
+1, & \text{ if } |p-i'| < |p-i| \\
-1, & \text{ otherwise}
\end{cases}
\end{equation}

In deep reinforcement learning, policy $\pi$ is represented by a deep neural network. Similar to the widely used policy networks for localization in 2D and 3D image space \cite{alansary2019evaluating, al2018partial}, our policy model also begins with a stack of trainable convolutional layers to extract discriminative features from the input state. These features are then fed to a fully connected multi-layer perceptron model with two output nodes for giving probabilities the two possible actions. Appendix~\ref{ap:policy} describes the policy network description in detail.

The goal of training is to optimize the policy model to maximize the episodic reward. Policy is usually optimized by stochastic gradient ascent over a number of epochs. At each epoch, the agent first runs several episodes using its current policy and gather experiences, where an experience $(\boldsymbol{s},a,\boldsymbol{s}',r)$ consists of the state $\boldsymbol{s}$, chosen action $a$, next state $\boldsymbol{s}'$, and the reward $r$. Sampling mini-batches from these experiences, the policy is then updated by taking gradient steps towards maximizing the expected reward. 

The gradients for the above-described policy-update are estimated following the policy gradient theorem \cite{mnih2016asynchronous}, which uses the log-likelihood of actions in its objective function. Despite the wide use of this objective function, it causes destructively large policy updates and is not robust \cite{schulman2017proximal, liu2019neural}. We use the recently proposed proximal policy optimization (PPO) \cite{schulman2017proximal}, which renders more robust and stable policy update. While in policy gradient theorem, log-probability is used to trace the impact of actions, PPO uses the ratio of the current policy to the old, i.e., $\rho(\pi) = \pi(\boldsymbol{s},a)/\pi_\text{old}(\boldsymbol{s},a)$. Furthermore, this impact is clipped between $[1-\delta, 1+\delta]$ in the objective function to prevent large policy changes. The clipped objective function is as follows:
\begin{equation}
\begin{split}
L(\pi) = \mathbb{E}_{(\boldsymbol{s},a,\boldsymbol{s}',r)} &\big[  \min \big(\rho(\pi)A(\boldsymbol{s},a,\boldsymbol{s}',r), \\ 
& \text{  clip}(\rho(\pi), 1-\delta, 1+\delta)A(\boldsymbol{s},a,\boldsymbol{s}',r)  \big) \big]
\end{split}
\label{eq:ppo_obj}
\end{equation} 
Here, $A(\boldsymbol{s},a,\boldsymbol{s}',r)$ indicates the advantage of the current policy for state $s$, over the previous state-value $V(\boldsymbol{s})$. State-value function $V(\boldsymbol{s})$ gives the discounted cumulative reward achievable through state $\boldsymbol{s}$ in the long-term. The discounted cumulative reward of the current policy for a transition $(\boldsymbol{s},a,\boldsymbol{s}',r)$ is estimated as: $r+\gamma V(\boldsymbol{s}')$, the discounted state-value of the next state added to the current reward. Here, $0<\gamma<1$ is the discount factor. Now, the advantage is computed by taking the increase of this cumulative reward from the value of state $\boldsymbol{s}$ : 
\begin{equation}
A(\boldsymbol{s},a,\boldsymbol{s}',r) = r+\gamma V(\boldsymbol{s}') - V(\boldsymbol{s})
\label{eq:discount}
\end{equation}

The state-value function $V$ is also represented by a deep neural network. Parameters of this value network is also updated at each epoch during training.

During testing with the trained policy $\pi$, the agent is initialized at a random position in the centerline depth world. Using the policy, it takes optimal actions to subsequently update its position until it moves back and forth at the orifice position. Setting the position of convergence in the centerline as the center of the orifice, we initialize a perpendicular plane as the orifice plane. Later, we refine this plane by tilting the plane upto $20$ degrees about each axis and finding the plane with minimum cross-section.

Using the centerline depth input, localizing the orifice in the centerline can also be done by other learning algorithms. For example, a 1D deep regressor can be used to directly estimate the orifice location in the centerline. Also, a spatial regressor could also be used to regress heatmaps around the orifice location. Though the search space is significantly reduced using centerline depth, the input dimension is still higher having a length equal to the centerline length. Collecting sufficient CT images to fit a deep model for this can be challenging. The RL-formulation, on the other hand, extracts states from the original centerline depth. These states are used as the training samples for policy training, which are much more in number gathered by agent-exploration. 

\section{Results and Discussion}
\begin{figure}
\includegraphics[scale=1.0]{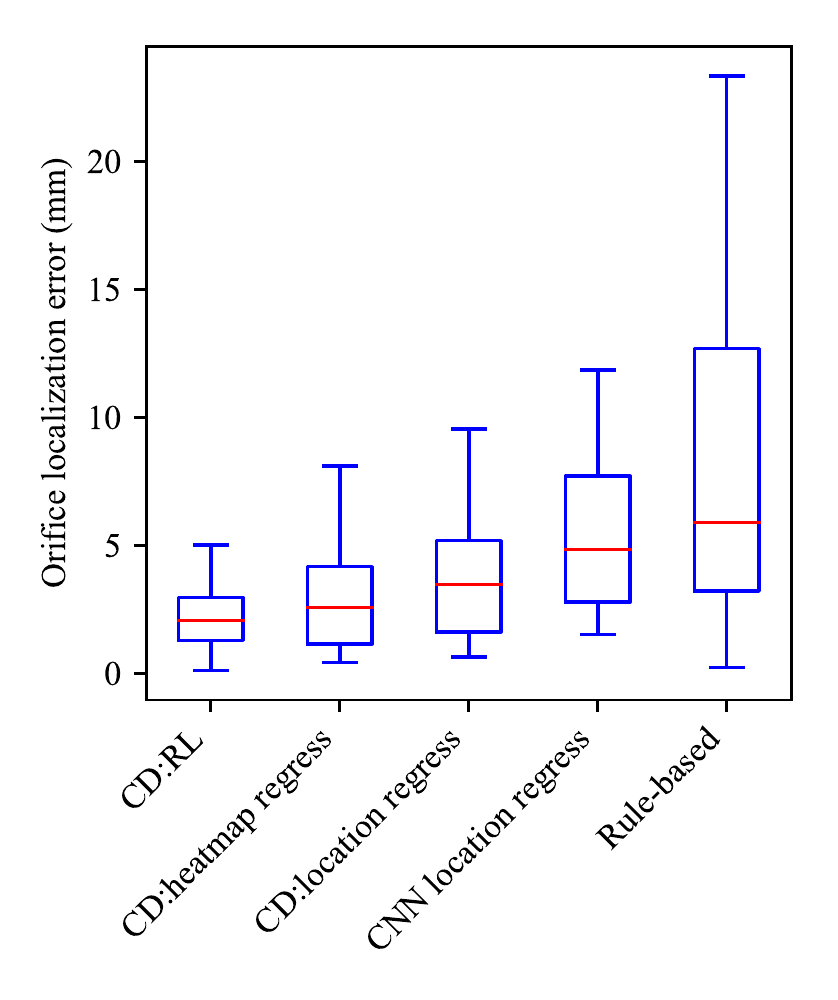}
\caption{{\bf Orifice localization error of different methods.} Previously attempted rule-based method showed a high error due to the sensitivity to the varied LAA anatomy in the CT images. End-to-end method using CNN also had high localization error, indicating the localization difficulty in full image space. Centerline depth-based methods showed relatively low error because of the reduced search space and simpler input feature. The proposed RL-based method could further lower the error by using its explorative learning with small states.}
\label{fig:boxplot}
\end{figure}

\begin{table*}
\caption{Orifice assessment error compared with expert annotation.}

\centering
\renewcommand{\arraystretch}{1.5}
\setlength{\tabcolsep}{6pt}
\begin{tabular}{c l c c c c c c}
\hline
& \multicolumn{1}{l}{\multirow{2}{*}{\bf Method}} 
& \multicolumn{2}{c}{\bf Center difference (mm)}
& \multicolumn{2}{c}{\bf Orientation difference ($^\circ$)}
& \multicolumn{2}{c}{\bf Area difference ($\text{mm}^2$)} \\
 & & Mean $\pm$ SD & Median
 & Mean $\pm$ SD & Median
 & Mean $\pm$ SD & Median\\
\hline
\parbox[t]{2mm}{\multirow{5}{*}{\rotatebox[origin=c]{90}{Overall}}}
 & {\bf CD:RL} & $\boldsymbol{ 2.547 \pm 1.931}$ & $\boldsymbol{2.067}$ & $\boldsymbol{ 9.895 \pm 6.236}$ & $\boldsymbol{9.982}$ & $\boldsymbol{ 72.399 \pm 74.365}$ & $\boldsymbol{44.165}$ \\ 
& CD:heatmap regress & $ 3.806 \pm 4.207$ & $2.582$ & $ 14.573 \pm 8.799$ & $12.354$ & $ 75.472 \pm 71.646$ & $54.029$ \\ 
& CD:location regress & $ 4.113 \pm 3.597$ & $3.485$ & $ 14.396 \pm 8.959$ & $13.581$ & $ 102.676 \pm 128.085$ & $54.821$ \\ 
& CNN location regress & $ 6.231 \pm 5.464$ & $4.851$ & $ 15.198 \pm 10.375$ & $12.194$ & $ 142.168 \pm 163.259$ & $102.160$ \\ 
& Rule-based & $ 8.904 \pm 8.058$ & $5.900$ & $ 18.213 \pm 12.208$ & $15.192$ & $ 355.463 \pm 534.821$ & $162.037$ \\
\hline
\parbox[t]{2mm}{\multirow{5}{*}{\rotatebox[origin=c]{90}{LAAO Patients}}}
& {\bf CD:RL} & $\boldsymbol{ 2.166 \pm 1.113}$ & $\boldsymbol{1.994}$ & $\boldsymbol{ 9.519 \pm 6.899}$ & $\boldsymbol{8.133}$ & $\boldsymbol{ 84.745 \pm 70.192}$ & $\boldsymbol{48.563}$ \\ 
& CD:heatmap regress & $ 4.075 \pm 6.193$ & $2.183$ & $ 15.282 \pm 11.762$ & $12.605$ & $ 121.863 \pm 186.677$ & $50.682$ \\ 
& CD:location regress & $ 3.539 \pm 1.744$ & $3.409$ & $ 10.887 \pm 8.076$ & $10.703$ & $ 127.723 \pm 148.023$ & $56.594$ \\ 
& CNN location regress & $ 6.709 \pm 4.329$ & $6.349$ & $ 17.542 \pm 13.335$ & $16.699$ & $ 304.937 \pm 308.462$ & $192.025$ \\ 
& Rule-based & $ 7.102 \pm 8.894$ & $5.322$ & $ 14.423 \pm 17.069$ & $9.584$ & $ 324.109 \pm 446.462$ & $168.534$ \\ 
\hline

\end{tabular}

\label{tab:loc}
\end{table*}

We evaluated the proposed orifice localization method using CT image of $98$ cases. $7$ cases among these are from actual LAAC patient who underwent through the appendage occlusion procedure using the WATCHMAN implant \cite{sick2007initial}. We randomly split the dataset to obtain $80\%$  of the data for training the policy. The remaining $20\%$ of the data is used as the test set for evaluating our localization model. During the split, the $7$ actual cases are kept in the test set. All the CT images are obtained from a local clinical site. Expert annotations for the orifice are also obtained from the same site. The pixel spacing of the volumes ranges from about $0.3$ to $0.5$ mm, where the inter-slice spacing ranges from $0.45$ to $0.50$ mm. The patients were about $67 \pm 8$ years old, about $36\%$ of them being female.

Because the previous orifice localization work \cite{leventic2019left} used a small dataset (of $17$ cases), we re-implemented their method using our data for fair comparison. In the result table and figures, we label this method as \emph{rule-based}. To show the advantage of the centerline-based search, we also compare the end-to-end deep localization models based on CNN to predict orifice location directly from CT image. We implemented both the direct location estimation network \cite{lv2017deep} and the heatmap regression network \cite{payer2019integrating}. However, they had slight difference in overall performance. Therefore, we only mention the results of direct location regression and identify this as \emph{CNN location regress} across tables and figures. 

We also implemented 1D versions of such models for the centerline-based search space, where the regression network directly predicts orifice location from the centerline depth input. This is useful for evaluating the necessity of using a RL agent to localize instead of direct regression from the centerline depth. All the centerline depth-based learning methods including ours are labeled with the prefix \emph{CD} indicating centerline depth.

We plot the localization error distribution of different methods in Fig.~\ref{fig:boxplot}. Here, the error represents the Euclidean distance of the localized orifice center from the expert annotated center location. For comprehensive evaluation, we also compare the orientation and area of the orifice in Table~\ref{tab:loc}. Orientation error is presented as the angular difference between the normal vectors of the detected and expert-identified orifice. In Table~\ref{tab:loc}, we also present the results for the actual LAAO patients separately.

\begin{figure*}[ht]
\centering
\includegraphics[scale=0.5]{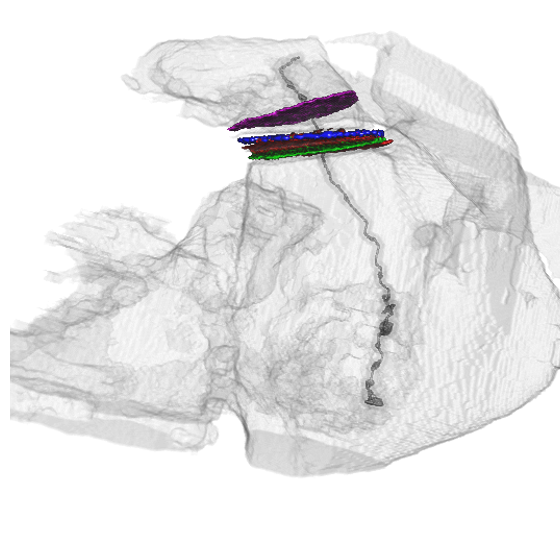}\hspace{25pt}\includegraphics[scale=0.5]{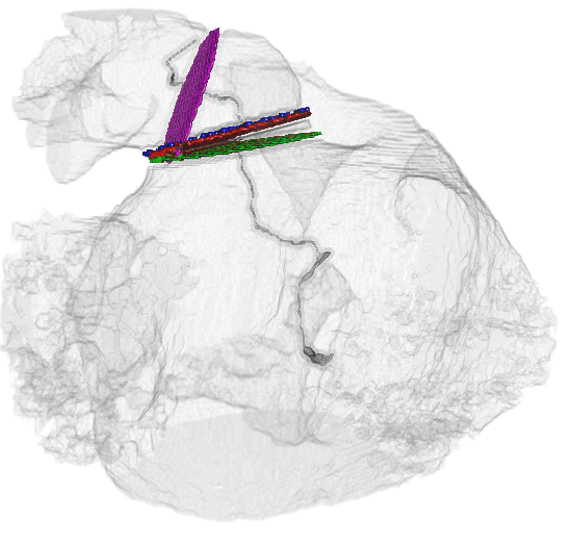}\hspace{25pt}\includegraphics[scale=0.5]{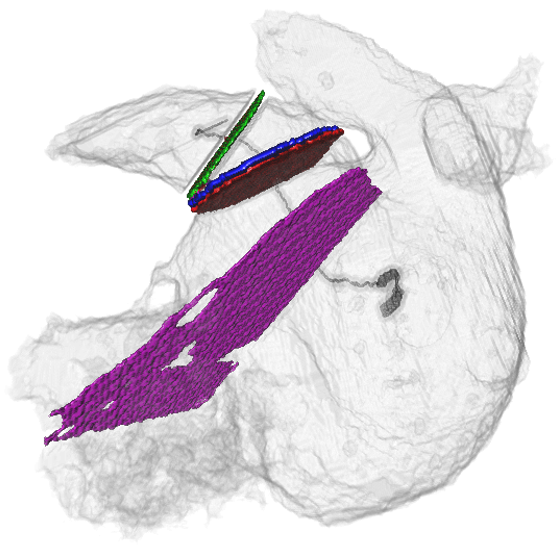}
\includegraphics[scale=0.84]{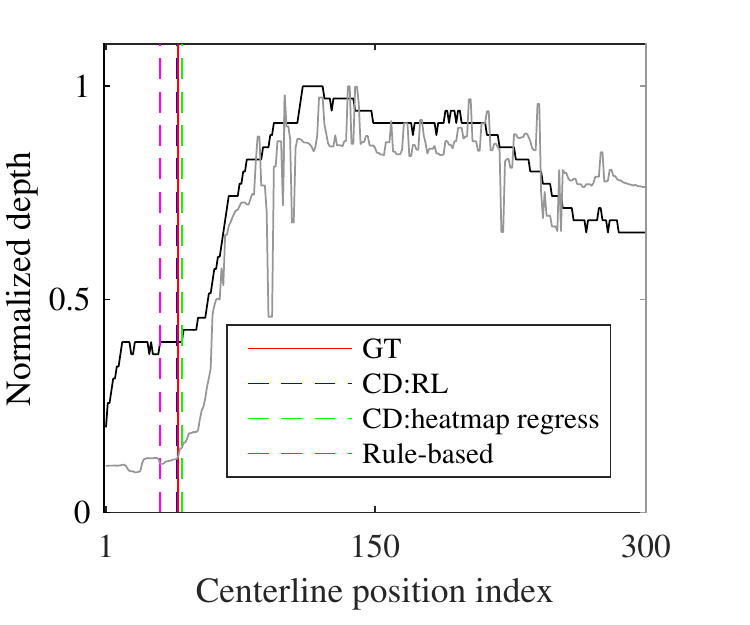}\hspace{-23pt}\includegraphics[scale=0.84]{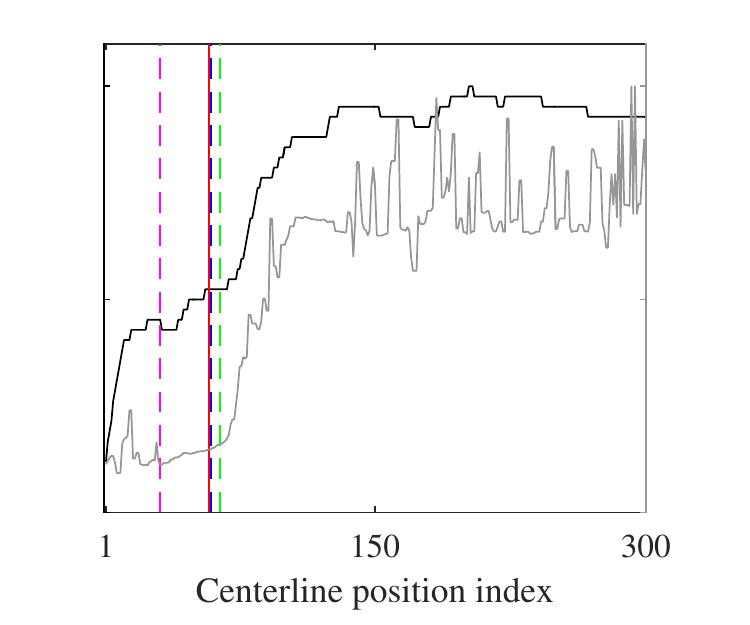}\hspace{-23pt}\includegraphics[scale=0.84]{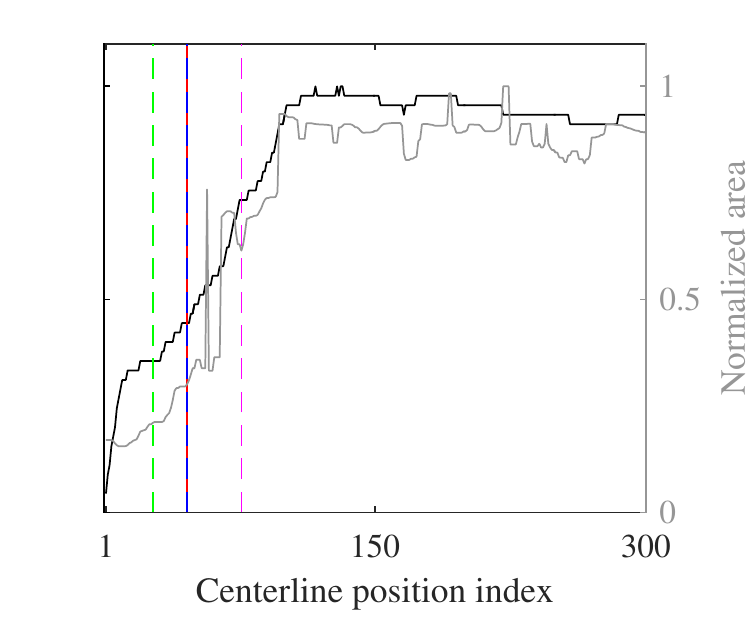}
\caption{{\bf Visualization of orifice location detected by different methods.} (Top) red, blue, green and magenta regions in the volumetric render indicate the resultant orifice by the expert, proposed method, centerline depth-based heatmap regression and previous rule-based approach, respectively. The black line is the centerline from LAA to LA. (Bottom) the resultant orifice locations are marked in the centerline depth map using the same colors. The proposed method could robustly localize the orifice near the expert-annotated location. The previous rule-based technique may fall in false local minima outside the neck (middle and right). RL-based method with its simpler model and small input states also resulted in better localization than the heatmap regression approach in cases where the enlargement starts much before the orifice. } 
\label{fig:cases}
\end{figure*}

The centerline depth-based localization methods could give significantly lower error than the end-to-end deep localization method. This shows the difficulty in localizing the small orifice using raw image with full search space. Limiting the search space to the centerline and using centerline depth input contributed to comparatively easier learning because it provides rich relatable information for tracking the anatomy. The previously proposed rule-based method \cite{leventic2019left}  showed relatively larger error. The irregular LAA structure affected the identification of the desired local minima before the largest uninterrupted cross-sectional increase. 

The proposed RL-based centerline search gave the lowest error with a reduced error variance. Having an average localization error of $2.547 \pm 1.931$ mm, the proposed method showed significant improvement over the other methods ($p$-value $<0.02$).
Centerline-based direct regression models used the small training data directly obtained from the given CT images. On the other hand, the proposed method utilizes the agent to explore thousands of sequential states for policy-training at each epoch. Moreover, the underlying state-distribution is also discovered by the agent as it attempts to perform different localization episodes through sequential state-transition. Thus, some states helpful for localization process are given more importance. Whereas, sample class-weight tuning is a crucial step in heatmap-regression based localization. 

 In Fig.~\ref{fig:cases}, we illustrate the localized orifice position for different cases. The first case shows an example where all the methods showed low localization error. In the second case, LAA neck is gradually enlarged from the tip. Therefore, the rule-based method gave incorrect localization assigning the local minima at the tip as the orifice. In the last case, there are multiple local minima within the enlargement from the orifice to the LA. As a result, the rule-based method falsely identified the orifice among such minima. Though the centerline-based deep localization model localized the orifice at a small distance in the previous two cases, it showed a high error for the last case. Because the tip is smaller than the neck, visually the anatomical enlargement starts earlier from the tip, such model proposed the orifice near the tip. Whereas, the actual orifice location falls within the enlargement, which is correctly localized by the proposed method.

\begin{figure*}[!ht]
\centering
\includegraphics[scale=1.0]{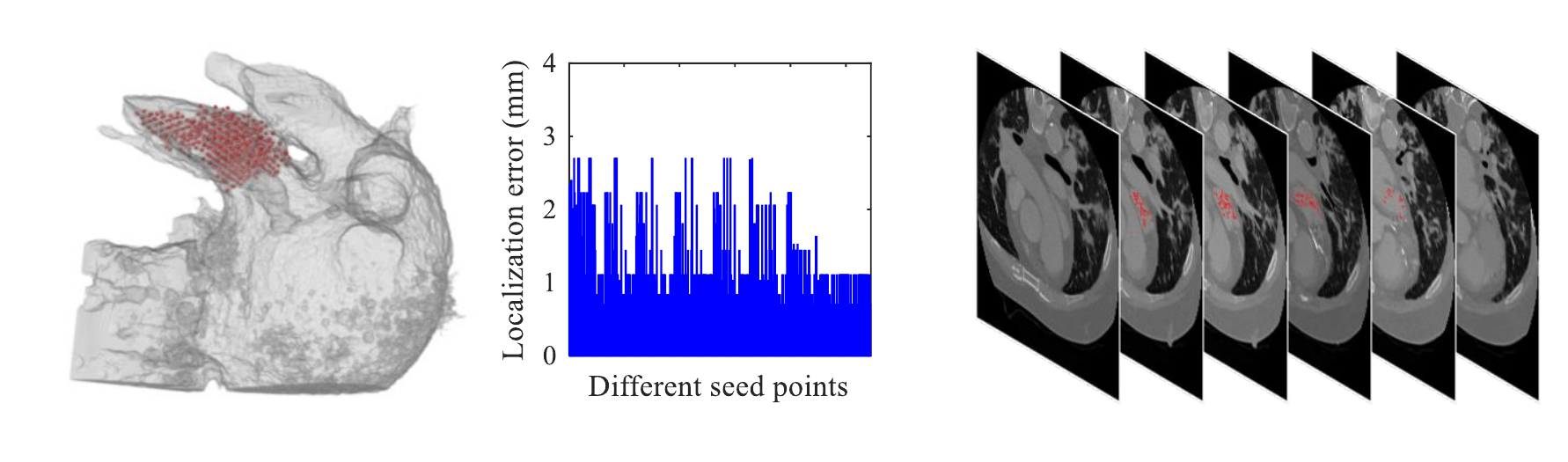}
\caption{{\bf Model sensitivity to seed initialization.} Detected orifice locations for different seed-points had a low variance. (Left) volumetric render of the seed-points inside cropped input volume. (Middle) Localization errors for different seed points. (Right) Seed points shown in the axial view.}
\label{fig:seed_test}
\end{figure*}

\begin{figure}[h]
\centering
\includegraphics[scale=1.0]{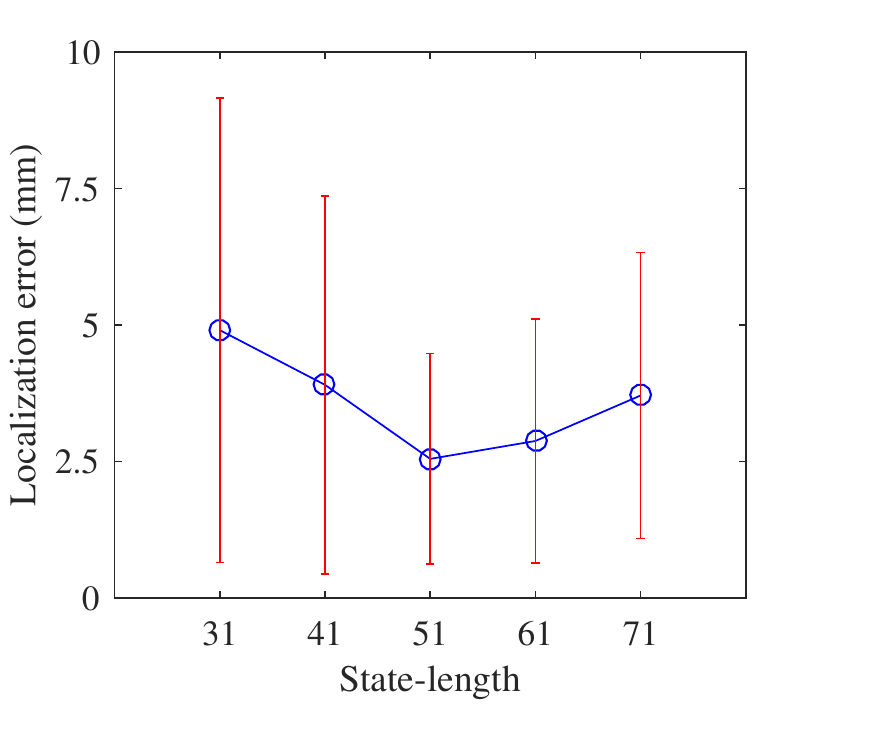}
\caption{{\bf Model dependency on state-length.} Too small states fail to provide sufficient information to guide the agent decision. Increasing the state-size improves the performance. Too big states reduce state-variation making the state-discrimination harder.}
\label{fig:state_length}
\end{figure}

To show the clinical impact of our method, we compared our localization results with the inter-observer variance. To do so, we obtained annotations for partial data from a second expert observer and compare these with the original annotations. The inter-observer difference of the orifice center was $2.82 \pm 1.42$ mm, which is larger than the difference of the proposed method. The whole localization process after manual seed initialization takes only about $8$ seconds, which is $18$ times faster than the previous approach \cite{leventic2019left}. Use of the linear centerline depth computation instead of exhaustive scanning for the minimum cross-sectional area contributed to such efficiency. Thus, the proposed method can be potentially useful for the physicians during the LAAO planning.

\subsection{Dependency on state-length}
State-based sequential modeling is the key reason to enable a simpler decision model with small but numerous training states. Therefore, state-length becomes an important parameter. We observed the localization performance of the proposed model by varying the state-size from $31$ to $71$. We plot the results in Fig.~\ref{fig:state_length}. Smaller states provide little information for the agent to decide optimal moves. Therefore, increasing the state-length allows the agent to utilize more information to effectively choose the action. However, after a certain state-length, the accuracy begins to degrade. Such bigger states reduces state-variation losing the local focus. Therefore, it becomes difficult for the agent to discriminate the states. In our experiment, a state-size of $51$ was optimal.

\subsection{Sensitivity to seed initialization}
Because our semi-automatic method requires manual seed initialization for growing the LAA centerline, analyzing the sensitivity to such initialization is necessary. Identifying the seed-point is rather simple because this does not have to be a precise location. Rather this can be anywhere in the LAA tip before the LAA neck. Though orifice identification is difficult in orthogonal CT views, LAA tip can be easily distinguished. Fig.~\ref{fig:seed_test}(right) shows various LAA tip seed-points in axial view. 

In Fig.~\ref{fig:seed_test}, we plot the localization performance for different seed-points from all over the LAA before the orifice. Resultant positions for different seed-points had a low variance of $0.302$. Therefore, manual annotation can be done with enough relaxation. From the annotated seed, we grow the centerline by tracking the local maxima in the depth map. Regardless of the initial position, the first few steps allows to reach the actual centerline (maximum depth nearby) crossing the highest gradient path in the depth map. Afterwards, the tracking follows along the desired centerline through the orifice upto the LA. Therefore, all such seed points eventually end up being grown into the actual centerline, causing similar localization results.

\section{Conclusion}
To ensure robust LAA orifice localization, we proposed a reinforcement learning-based approach. We formulated a centerline depth world where a localization agent performs a search for the orifice on the LAA centerline observing local centerline depth as small states. The proposed approach showed improved robustness compared with the previously rule-based approach. It also showed significant improvement over the end-to-end CNN-based localization, by using a reduced search space and simpler input. Moreover, comparing deep learning-based methods using the same reduced search space, we observed that the proposed RL formulation 
enabled more effective learning by its sequential exploration of small training states. Therefore, the proposed localization approach can be a useful aid to the physicians during the LAA occlusion planing.

\appendices
\section{Policy Network and RL hyperparameters}
\label{ap:policy}
The initial part of the policy network has three 1D convolutional units to output high-level feature from the input state. Each unit has two convolutional layers with kernel-size of $3$, using rectified linear unit (ReLU) activation. At the end of each unit, there is a 1D max-pooling layer having a pool-size of $2$ and stride of $2$ to reduce the spatial dimension. 
The latter part of the network consists of three fully connected (FC) layers. The final FC layer has two output nodes and uses softmax-gating to produce probabilities for the actions in our binary action space.  

During the training, an $\epsilon$-greedy policy is used for episodic state-exploration, where the agent greedily chooses an action based on the policy with probability $\epsilon$ and it chooses a random action with probability $1-\epsilon$. In our experiment, $\epsilon=0.7$ is used. The discount factor $\gamma$ in \eqref{eq:discount} is set to $0.9$. Both these parameter-values are widely used in different RL studies \cite{alansary2019evaluating, al2018partial}. The optimal learning rate was chosen to be $1\mathrm{e}{-5}$, after trial-and-error process using different learning rates ranging from $0.1$ to $1\mathrm{e}{-7}$.

\bibliography{mybib}

\EOD

\end{document}